\documentclass[conference]{IEEEtran}
\IEEEoverridecommandlockouts
\usepackage{placeins} 
\usepackage{cite}
\usepackage{amsmath,amssymb,amsfonts}
\usepackage{algorithmic}
\usepackage{graphicx}
\usepackage{textcomp}
\usepackage{xcolor}
\usepackage{hyperref}
\usepackage{comment}
\usepackage[natbibapa]{apacite} 
\def\BibTeX{{\rm B\kern-.05em{\sc i\kern-.025em b}\kern-.08em
    T\kern-.1667em\lower.7ex\hbox{E}\kern-.125emX}}

\begin{document}

\title{Emoji Prediction in Tweets using BERT\\

}

\author{\IEEEauthorblockN{\textsuperscript{} Muhammad Osama Nusrat*}
\IEEEauthorblockA{\textit{Department of Computing} \\
\textit{Fast Nuces}\\
Islamabad, Pakistan \\
i212169@nu.edu.pk}
\and
\IEEEauthorblockN{\textsuperscript{} Zeeshan Habib}
\IEEEauthorblockA{\textit{Department of Computing} \\
\textit{Fast Nuces}\\
Islamabad, Pakistan \\
i212193@nu.edu.pk}

\and
\IEEEauthorblockN{\textsuperscript{} Mehreen Alam}
\IEEEauthorblockA{\textit{Department of Computing} \\
\textit{Fast Nuces}\\
Islamabad, Pakistan \\
mehreen.alam@nu.edu.pk}

\and
\IEEEauthorblockN{\textsuperscript{}  Saad Ahmed Jamal}
\IEEEauthorblockA{\textit{Department of Computer Science} \\
\textit{Université Bretagne Sud}\\
Vannes, France \\
jamal.e2107235@etud.univ-ubs.fr
}

}

\maketitle

\begin{abstract}

In recent years, the use of emojis in social media has increased dramatically, making them an important element in understanding online communication. However, predicting the meaning of emojis in a given text is a challenging task due to their ambiguous nature. In this study, we propose a transformer-based approach for emoji prediction using BERT, a widely-used pre-trained language model. We fine-tuned BERT on a large corpus of text (tweets) containing both text and emojis to predict the most appropriate emoji for a given text. Our experimental results demonstrate that our approach outperforms several state-of-the-art models in predicting emojis with an accuracy of over 75 percent. This work has potential applications in natural language processing, sentiment analysis, and social media marketing.

\end{abstract}

\section{Introduction}

In the past few years, social media has emerged as a prolific source of data for numerous research fields, with natural language processing (NLP) being one of them. As a result of the widespread use of mobile devices and the internet, social media platforms such as Twitter have become a popular means for people to express their emotions, opinions, and sentiments on various topics. In this context, emojis have become a popular way of conveying emotions and sentiments in text-based communication. Emojis are small pictograms that represent emotions, objects, or concepts and are widely used on social media platforms. 

Emoji prediction is a task that involves predicting the most appropriate emoji to use in a given textual conversation based on the context of the conversation. This task is essential in improving the effectiveness of communication on social media platforms, especially in situations where the text is ambiguous, and the use of emojis can add clarity to the message.

To address the challenge of emoji prediction, recent studies have explored the use of transformer models, particularly the Bidirectional Encoder Representations from Transformers (BERT) model. BERT is a powerful pre-trained transformer model that has shown state-of-the-art performance in a wide range of natural language processing tasks.

The use of BERT in emoji prediction involves fine-tuning the model on a large dataset of tweets or other social media posts to learn the contextual relationships between the text and the appropriate emojis. The fine-tuned model can then be used to predict the most appropriate emoji to use in a given context.

Despite the promising results reported by recent studies on emoji prediction using transformer models, there are still some challenges that need to be addressed. One of the challenges is the lack of large, diverse datasets for training and evaluating the models. Another challenge is the diversity of emojis used in different languages and cultures, which requires the development of language-specific and culture-specific models.

In this context, this study explores the use of BERT for emoji prediction in a dataset of tweets. We fine-tune the BERT model on a large dataset of tweets and evaluated its performance on a test set of tweets. We also examined the impact of different factors, such as the size of the training data and the number of emojis, on the performance of the model. The findings of this study provided insights into the effectiveness of transformer models for emoji prediction and can contribute to the development of more accurate and efficient emoji prediction models for social media platforms.

\section{Literature Review}

The authors present a groundbreaking approach to pre-train language models that that has since become one of the most influential contributions to NLP in recent years \citep{DBLP:journals/corr/abs-1810-04805} . They proposed an approach called BERT, Bidirectional Encoder Representations from Transformers, is a deep learning architecture that uses a bidirectional transformer network to pre-train a language model on a large amount of unlabelled datasets. The model is then fine-tuned on some NLP tasks like text classification or question answering. They have described their approach to pretraining BERT, including the use of a novel masked language modeling objective that randomly masks tokens in the input sequence and then model predict the masked tokens based on the surrounding context. This objective allows BERT to capture both local and global context in the input sequence, resulting in a highly contextualized representation of language. The authors also describe their use of a next-sentence prediction objective, which helps BERT capture the relationship between two sentences in a document.

The author argue that language models, which are traditionally trained to predict the next word in a sentence or the likelihood of a sentence given a context, can be viewed as multitask learners that can perform a variety of tasks without explicit supervision \citep{Radford2019LanguageMA}. They propose a method for training language models on a diverse set of tasks, including sentiment analysis, question answering, and language translation, without any labelled data. The model is trained on a new dataset of millions of webpages called WebText. The approach, called Unsupervised Multi-task Learning (UMT), manipulating the vast amounts of unannotated text available on the internet to train a single neural network on multiple tasks simultaneously. By sharing the parameters across tasks, the model is able to learn from the common underlying structure of language and perform well on a range of tasks. The authors also introduce a new benchmark, called the General Language Understanding Evaluation (GLUE), which measures the performance of language models on a suite of diverse NLP tasks. Using UMT, they achieve state-of-the-art results on the GLUE benchmark, outperforming previous approaches that relied on supervised learning.

\citet{DBLP:journals/corr/abs-2109-01652} proposed a new approach to enhance the zero-shot learning ability of language models by combining the pre-training and fine-tuning paradigm with prompting. Their method involves fine-tuning a pre-trained model with 137 billion parameters on a range of datasets described through instructions. By evaluating the model's performance on previously unseen tasks, the authors demonstrated that their instruction-tuned model, FLAN (Finetuned Language Net), outperformed its untuned counterpart by a significant margin in a zero-shot setting. Additionally, FLAN surpassed GPT-3 in zero-shot performance on 20 out of 25 datasets evaluated, indicating its superior performance.

\citet{Felbo_2017} proposed a novel sentiment, emotion, and sarcasm detection approach using millions of emoji occurrences as a weakly supervised learning signal. The authors introduce the DeepMoji model, a deep learning architecture based on long short-term memory (LSTM) networks. The model is pre-trained on a large dataset containing 1.2 billion tweets with emoji, allowing it to learn semantic representations of text from these noisy labels. This pre-training approach helps learn effective representations for downstream tasks such as sentiment analysis, emotion recognition, and sarcasm detection. The DeepMoji model demonstrates state-of-the-art performance on several benchmarks, outperforming existing methods. This work highlights the potential of using emojis as a weak supervision signal to learn domain-agnostic representations that can be effectively used for various natural language processing tasks.

\citet{ma2020emoji} build upon the work of \citet{Felbo_2017} by exploring the problem of emoji prediction more comprehensively. The authors introduce several extensions to the DeepMoji model, such as incorporating attention mechanisms, leveraging tweet metadata, and utilizing pre-trained language models like BERT. The authors also present a new benchmark dataset called 'EmoBank', which is collected from Twitter and contains 4.7 million tweets with emoji. EmoBank is designed to evaluate models on various emoji prediction tasks, such as predicting the presence, absence, and type of emojis in a given text. The extended model shows improved performance compared to the original DeepMoji model and other baselines, demonstrating the effectiveness of the proposed extensions.

\citet{wolf2019huggingface} proposed a novel neural network architecture called the Transformer, which relies solely on self-attention mechanisms, discarding the need for traditional recurrent or convolutional layers. The authors argue that attention mechanisms can model long-range dependencies and parallel computation more effectively than LSTMs or CNNs, thus addressing some of the limitations of these traditional architectures. The Transformer model achieves state-of-the-art results on various natural language processing tasks, including machine translation and language modeling. This work has significantly impacted the field, inspiring a range of follow-up research and developing powerful pre-trained language models such as BERT and GPT-2.

\citet{brown2020language} introduced a new language model called GPT 3 which was an advancement to GPT 2 as it solved some of the problems which were addressed in the previous language model. GPT-2 required rigorous fine tuning to do a specific task. GPT-3 solved this problem as it does not require a lot of fine tuning to do a particular task such as in language translation, GPT 3 can translate a sentence from one language to another with just a few samples whereas in GPT-2 we had to provide relatively more samples so that the model perform well. Similarly GPT-3 outperforms GPT-2 in question answering, filling missing words in a sentence, using new words in the sentence which are not present in the vocabulary, doing calculations and many other tasks. We can confidently say GPT-3 is a better short learner than GPT-2. By few shot, we mean the ability to learn with few examples. The reason for the success of GPT-3 is that it has more parameters than GPT-2. GPT-3 contains 175 billion massive parameters compared to GPT-2 which has only 1.5 billion parameters. GPT-3 has brought ease in many NLP domains where we had very less labeled data and it was nearly impossible to get results with such small labeled examples. GPT-3 has made it solvable now. For e.g we can build a chatbot for a travel agency with very few examples with GPT-3 whereas previously, we required huge amounts of labeled data to do the same task. The author also highlighted some limitations of the GPT-3 model which included that GPT-3 does not understand the context of the document properly for e.g if we ask it to write a summary of a scientific paper it will fail to capture all important points and write a summary. Moreover if we ask it to generate a response for a complaint it may output a random response which may not address the user's problem. GPT-3 also has another limitation as it generates biased outputs because it is trained on a data which is more male biased. For e.g it may write negatively about woman such as woman are not suitable for leadership positions and woman cannot drive safe etc which is not okay.

\citet{NIPS2017_3f5ee243} discussed how transformers have revolutionized natural language processing tasks. Transformers have enabled machines to generate human-like content. Transformer architecture was introduced in 2017 in the famous paper 'ATTENTION IS ALL YOU NEED'. It solved several previous issues addressed in RNN, such as bottleneck problems and long-range dependency issue problems. RNNs cannot capture information when sentences are long due to vanishing gradient issues. Transformer solved this problem because it is based on a self-attention mechanism focusing on the sentence's essential parts. Moreover, the transformer uses multi-head attention, which means it consists of multiple attention mechanism which helps to focus on multiple parts of the input sentence in parallel. Each head focuses on different parts of the input sentence. One head can focus on the sentence's subject, the other on an object, and the third on the object. In multi-head attention, instead of a single context vector, multiple context vectors are generated, which contain the input sentence information, which results in a better performance than when we use a single attention mechanism. Moreover, transformers are faster than recurrent neural networks as they can handle parallel processing. We can use transformers to do many tasks by fine-tuning them on small datasets, as it has been trained on large datasets. Transformers use an attention mechanism which makes it great for summarizing articles and research papers because it focuses on essential parts of the documents and then gathers all those critical points to generate a summary.  The author then introduced a library called TRANSFORMERS which is an open-source library that students and scientists can use to do NLP tasks more efficiently. Primary purpose of building this library was to provide ease to the people. Instead of writing code from scratch, they can use this library which would save time and energy. This library can be used to do multiple NLP tasks such as sentiment analysis, text classification, question answering, and language generation. The library contains many pre-trained models, such as BERT. GPT-2. RoBERTa, DistilBERT, T5 etc. These models have been trained on a massive amount of text data, such as Wikipedia, and these pre-trained models can then be fine-tuned to our task. means These pretrained models requires less training time and fewer data instead of training from scratch.

\section{Methodology}

\subsection{Pipeline}

A methodological outline, the process for developing the natural language processing (NLP) model using a
neural network is shown in figure \ref{fig:you1}. The model was built using keras python library. 
In the following, we will go through each step:

\begin{figure}[!ht]
    \begin{center}
        \includegraphics[width=09cm]{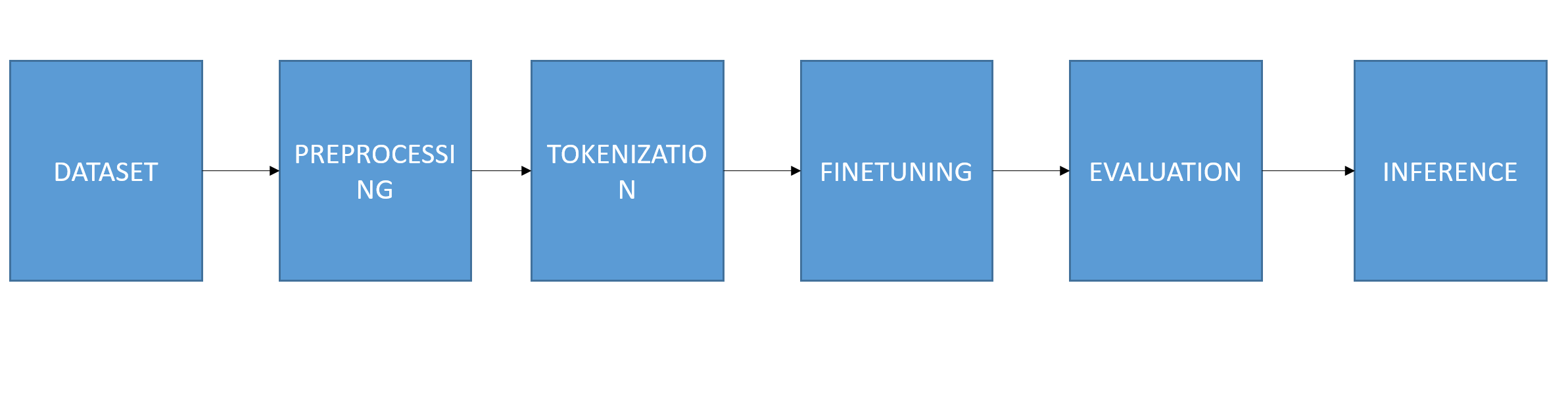}
        \caption{Pipeline}
        \label{fig:you1}
    \end{center}
\end{figure}

Dataset collection
was the initial step, where data was gathered, needed to train and evaluate the NLP model. The dataset that relevant was scrutinized and it contained a representative sample of the text and emojis. The quality of collected dataset directly influences the performance and generalization ability of deep learning model.

Raw text data that scrubbed contained noise, inconsistencies, and irrelevant information. Preprocessing steps included cleaning, stemming and transforming the data to make it suitable for training. This included tasks like removing punctuation, converting text to lowercase, handling special characters, and dealing with missing values.
It was one of the most time consuming task. Stemming is a natural language processing (NLP) technique used to reduce words to their base or root form, which may not be a valid word itself but can still convey the word's essential meaning [10]. The process involves removing prefixes, suffixes, and other inflections from words to obtain the core linguistic stem. In Python, the Natural Language Toolkit (NLTK) and other libraries provide various stemming algorithms. For this task, NLTK was used. Also, the impact of using stemming was assessed on the results.

Tokenization and Embedding involved breaking down the text into smaller units called tokens and  to represent words as numerical vectors in a high-dimensional space. These vectors capture semantic relationships and contextual information between words, allowing machine learning models to better understand and process textual data. In this context, tokens were usually words. It is a crucial step because it converts text data into a format that a neural network can understand and process. The result of this step is represented as a unique integer index array.

Finetuning included the process of training a pre-existing neural network model on your specific task or dataset. The BERT model pretrained on large language corpora to learn general language features was finetuned on our dataset. Finetuning involved updating the model's parameters to make it more specialized and accurate for predicting multiple emojis.

After finetuning the model, the performance of model was accessed through a set of evaluation metrics on train, validation and test sets. The model was trained on the training set and then evaluated on the test set. 
Once the model was trained and evaluated, it was used for making predictions on new, unseen data. Inference included to the process of feeding new text inputs to the trained model and obtaining predictions.

\subsection{Dataset}

The first dataset containing tweets and label emojis were stored in a csv file. It was a small dataset comprising of 132 rows for training and 56 rows for testing. There were 5 emoji classes in the dataset. The train test split ratio was selected as 70:30. 

The second dataset has 2 csv files, Train and Test. Train contained tweets with the emoji label in coded form. There were 69,832 tweets in train while the test had 25,920 tweets. In-addition, there were two supplementary csv file Mapping and Output. Mapping file contained emojis with their label mapping. The fourth csv file output, contained unique ids. There were 20 emoji classes in this dataset while the train test split ratio was kept as same. 

The smaller dataset 1 served as initial setup for the model before actual training on larger dataset 2. After the initial setup and some level of learning from the smaller dataset, the model was then further trained using a larger and more comprehensive dataset. This second phase of training, using the larger dataset, aimed to improve the model's performance and accuracy by exposing it to a wider range of data.

\subsection{Evaluation Metric}
Training and Validation loss were monitored while the model was being trained.
For testing,
we have used F1 score as our evaluation metric while keeping in view precision, accuracy, recall values as well. 
Following is the formula for the metric for F1 Score.

\begin{equation}
    F_1Score = 2 * \frac{(Precision * Recall)}  {(Precision + Recall)}
\end{equation}

\subsection{Approach}

BERT is a highly advanced pre-trained language model developed by Google that uses a bidirectional approach and deep neural network to better understand natural language by analyzing the entire input sequence in both directions during training, leading to more accurate language processing and understanding. It has been widely used in various natural language processing tasks, improving the accuracy and effectiveness of NLP applications and inspiring the development of other advanced pre-trained language models \citep{DBLP:journals/corr/abs-1810-04805}.
BERT has several advantages, making it a popular choice for natural language processing tasks. Firstly, it uses a bidirectional approach during training, which allows it to better understand the context of words in a sentence. This can lead to more accurate language processing and understanding than other language models that only process text in one direction.
Secondly, BERT has been pre-trained on a large corpus of text data, allowing it to capture many language patterns and nuances. This makes it highly effective for various NLP tasks, such as question answering, sentiment analysis, and language translation.
Several other pretrained models, designed for different tasks, were tried to check the performance including the model used by \citet{jamal2023data} for multi task learning and \citet{app10196878} for hydrological modeling. Finally, BERT has inspired the development of other advanced pre-trained language models, such as GPT-3 and RoBERTa. These models build on BERT's success and improve its architecture, resulting in even better performance in natural language processing tasks.

\section{Results \& Discussion}

The deep learning model contained a BERT pretrained model and a dense network. For the dense network, different settings for number of hidden layers and neurons were experimented. The characteristics of selected model for the task included three dense layers with pooling layers.

\begin{figure}[!ht]
    \begin{center}
        \includegraphics[width=4.8cm]{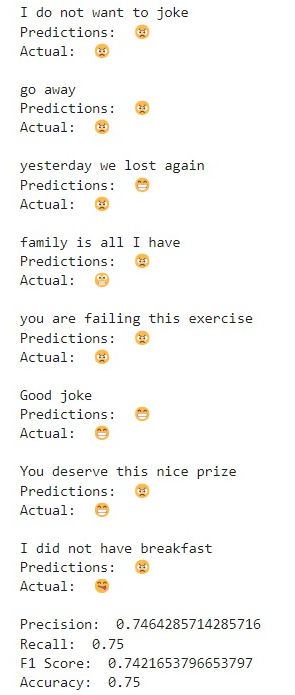}
        \caption{Actual output vs predicted output}
        \label{fig:your-label2}
    \end{center}
\end{figure}

The fine tuned model was used to predict emojis on the unseen data. Figure \ref{fig:your-label2}  shows a snapshot that the model correctly predicts the emojis corresponding to the tweets of test dataset.
Precision, recall, F1 score, and accuracy were observed as 75.4\%, 73.2\%, 73\%, and 73.2\%, respectively on dataset 1.
Similarly on dataset 2 the values of precision,recall, F1 score, and accuracy were
40\%, 60\%, 46\%, and 60\%, respectively.
The model was trained for 10 epochs for dataset 1 and 800 epochs for dataset 2. The results suggest that on dataset 2, the model's precision was relatively low at 40\%, indicating that it made a significant number of false positive predictions. However, it achieved a recall value is indicating that it captured a reasonable portion of actual positive instances. 
These results provide insights into how well the model is performing in terms of both false positives and false negatives in dataset 2.

\begin{figure*}
    \begin{center}
    \includegraphics[width=15cm]{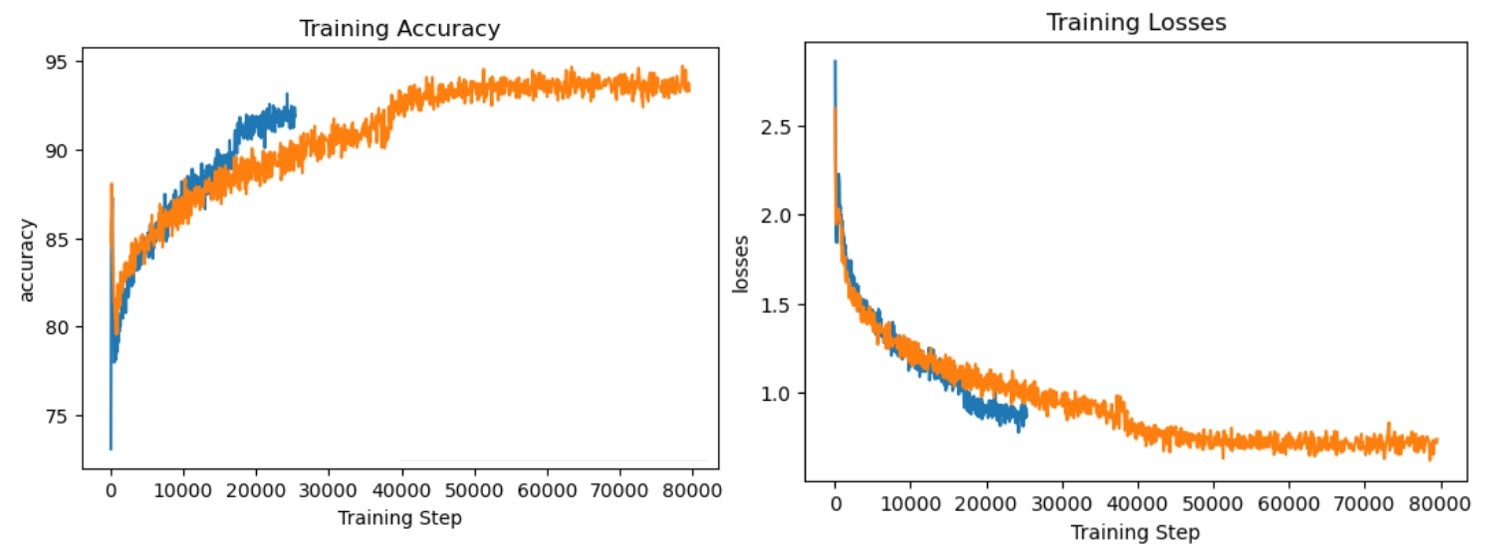}
    \caption{Accuracy and Losses, orange curve represents training Loss while blue curve represents validation loss}
    \label{fig:galaxy4_ari}
    \end{center}
\end{figure*}

The training curve as in figure \ref{fig:galaxy4_ari} displays the progression of the training loss as the BERT model was iteratively trained over a certain number of epochs. As depicted, the training loss exhibits an initial fluctuation during the early epochs, which is indicative of the model's exposure to diverse patterns within the training data.

Subsequently, the training loss consistently decreases, indicative of the model effectively capturing the underlying structures and representations within the data. This phase of diminishing loss continues until convergence is reached, where further reductions in loss become marginal. The smooth convergence demonstrates the BERT model's ability to learn intricate linguistic features from the training data.
The training and validation loss decreased as the number of epochs increased and the accuracy increased with increase in the number of epochs. 
This effect can be seen in figure \ref{fig:galaxy4_ari}.

\FloatBarrier
\section{Conclusion}

In conclusion, this study demonstrated the effectiveness of using BERT for emoji prediction on a dataset of tweets. We found that with fine-tuning a pre-trained BERT model on a dataset of labeled tweets, state-of-the-art results can be achieved on this task. Our experiments showed that the BERT- model based approach outperforms traditional machine learning models and other deep learning models. Additionally, our study highlights the importance of pre-processing techniques such as tokenization and stemming for improving model performance. Furthermore, it was found that using tweet-specific features such as hashtags and user mentions as input features can further improve model performance. Results suggest that BERT can be a valuable tool for predicting emojis in tweets, which can be useful for a variety of applications such as social media monitoring and sentiment analysis.

\section*{Dataset and Code Availability}
The datasets used for this research are available on kaggle through the following link
\href{https://www.kaggle.com/datasets/alvinrindra/emojify}{https://www.kaggle.com/datasets/alvinrindra/emojify,}
\href{https://www.kaggle.com/code/sakthinarayanan/emoji-predictor/input}{https://www.kaggle.com/code/sakthinarayanan/emoji-predictor/input.}
The developed code is made available at Github: \href{https://github.com/mnusrat786/emoji-prediction-with-transformer}{https://github.com/mnusrat786/emoji-prediction-with-transformer}

\bibliography{mybibliography}

\begin{thebibliography}{}

\bibitem [\protect \citeauthoryear {%
Brown%
\ \protect \BOthers {.}}{%
Brown%
\ \protect \BOthers {.}}{%
{\protect \APACyear {2020}}%
}]{%
brown2020language}
\APACinsertmetastar {%
brown2020language}%
\begin{APACrefauthors}%
Brown, T\BPBI B.%
, Mann, B.%
, Ryder, N.%
, Subbiah, M.%
, Kaplan, J.%
, Dhariwal, P.%
\BDBL {}Amodei, D.%
\end{APACrefauthors}%
\unskip\
\newblock
\APACrefYearMonthDay{2020}{}{}.
\newblock
\APACrefbtitle {Language Models are Few-Shot Learners.} {Language models are
  few-shot learners.}
\PrintBackRefs{\CurrentBib}

\bibitem [\protect \citeauthoryear {%
Devlin%
, Chang%
, Lee%
\BCBL {}\ \BBA {} Toutanova%
}{%
Devlin%
\ \protect \BOthers {.}}{%
{\protect \APACyear {2018}}%
}]{%
DBLP:journals/corr/abs-1810-04805}
\APACinsertmetastar {%
DBLP:journals/corr/abs-1810-04805}%
\begin{APACrefauthors}%
Devlin, J.%
, Chang, M.%
, Lee, K.%
\BCBL {}\ \BBA {} Toutanova, K.%
\end{APACrefauthors}%
\unskip\
\newblock
\APACrefYearMonthDay{2018}{}{}.
\newblock
{\BBOQ}\APACrefatitle {{BERT:} Pre-training of Deep Bidirectional Transformers
  for Language Understanding} {{BERT:} pre-training of deep bidirectional
  transformers for language understanding}.{\BBCQ}
\newblock
\APACjournalVolNumPages{CoRR}{abs/1810.04805}{}{}.
\newblock
\begin{APACrefURL} \url{http://arxiv.org/abs/1810.04805} \end{APACrefURL}
\PrintBackRefs{\CurrentBib}

\bibitem [\protect \citeauthoryear {%
Felbo%
, Mislove%
, S{\o}gaard%
, Rahwan%
\BCBL {}\ \BBA {} Lehmann%
}{%
Felbo%
\ \protect \BOthers {.}}{%
{\protect \APACyear {2017}}%
}]{%
Felbo_2017}
\APACinsertmetastar {%
Felbo_2017}%
\begin{APACrefauthors}%
Felbo, B.%
, Mislove, A.%
, S{\o}gaard, A.%
, Rahwan, I.%
\BCBL {}\ \BBA {} Lehmann, S.%
\end{APACrefauthors}%
\unskip\
\newblock
\APACrefYearMonthDay{2017}{}{}.
\newblock
{\BBOQ}\APACrefatitle {Using millions of emoji occurrences to learn any-domain
  representations for detecting sentiment, emotion and sarcasm} {Using millions
  of emoji occurrences to learn any-domain representations for detecting
  sentiment, emotion and sarcasm}.{\BBCQ}
\newblock
\BIn{} \APACrefbtitle {Proceedings of the 2017 Conference on Empirical Methods
  in Natural Language Processing.} {Proceedings of the 2017 conference on
  empirical methods in natural language processing.}
\newblock
\APACaddressPublisher{}{Association for Computational Linguistics}.
\newblock
\begin{APACrefURL} \url{https://doi.org/10.18653%2Fv1%2Fd17-1169}
  \end{APACrefURL}
\newblock
\begin{APACrefDOI} \doi{10.18653/v1/d17-1169} \end{APACrefDOI}
\PrintBackRefs{\CurrentBib}

\bibitem [\protect \citeauthoryear {%
Jamal%
\ \BBA {} Aribisala%
}{%
Jamal%
\ \BBA {} Aribisala%
}{%
{\protect \APACyear {2023}}%
}]{%
jamal2023data}
\APACinsertmetastar {%
jamal2023data}%
\begin{APACrefauthors}%
Jamal, S\BPBI A.%
\BCBT {}\ \BBA {} Aribisala, A.%
\end{APACrefauthors}%
\unskip\
\newblock
\APACrefYearMonthDay{2023}{}{}.
\newblock
\APACrefbtitle {Data Fusion for Multi-Task Learning of Building Extraction and
  Height Estimation.} {Data fusion for multi-task learning of building
  extraction and height estimation.}
\PrintBackRefs{\CurrentBib}

\bibitem [\protect \citeauthoryear {%
Ma%
, Liu%
, Wang%
\BCBL {}\ \BBA {} Vosoughi%
}{%
Ma%
\ \protect \BOthers {.}}{%
{\protect \APACyear {2020}}%
}]{%
ma2020emoji}
\APACinsertmetastar {%
ma2020emoji}%
\begin{APACrefauthors}%
Ma, W.%
, Liu, R.%
, Wang, L.%
\BCBL {}\ \BBA {} Vosoughi, S.%
\end{APACrefauthors}%
\unskip\
\newblock
\APACrefYearMonthDay{2020}{}{}.
\newblock
\APACrefbtitle {Emoji Prediction: Extensions and Benchmarking.} {Emoji
  prediction: Extensions and benchmarking.}
\PrintBackRefs{\CurrentBib}

\bibitem [\protect \citeauthoryear {%
Nusrat%
\ \protect \BOthers {.}}{%
Nusrat%
\ \protect \BOthers {.}}{%
{\protect \APACyear {2020}}%
}]{%
app10196878}
\APACinsertmetastar {%
app10196878}%
\begin{APACrefauthors}%
Nusrat, A.%
, Gabriel, H\BPBI F.%
, Haider, S.%
, Ahmad, S.%
, Shahid, M.%
\BCBL {}\ \BBA {} Ahmed~Jamal, S.%
\end{APACrefauthors}%
\unskip\
\newblock
\APACrefYearMonthDay{2020}{}{}.
\newblock
{\BBOQ}\APACrefatitle {Application of Machine Learning Techniques to Delineate
  Homogeneous Climate Zones in River Basins of Pakistan for Hydro-Climatic
  Change Impact Studies} {Application of machine learning techniques to
  delineate homogeneous climate zones in river basins of pakistan for
  hydro-climatic change impact studies}.{\BBCQ}
\newblock
\APACjournalVolNumPages{Applied Sciences}{10}{19}{}.
\newblock
\begin{APACrefURL} \url{https://www.mdpi.com/2076-3417/10/19/6878}
  \end{APACrefURL}
\newblock
\begin{APACrefDOI} \doi{10.3390/app10196878} \end{APACrefDOI}
\PrintBackRefs{\CurrentBib}

\bibitem [\protect \citeauthoryear {%
Radford%
\ \protect \BOthers {.}}{%
Radford%
\ \protect \BOthers {.}}{%
{\protect \APACyear {2019}}%
}]{%
Radford2019LanguageMA}
\APACinsertmetastar {%
Radford2019LanguageMA}%
\begin{APACrefauthors}%
Radford, A.%
, Wu, J.%
, Child, R.%
, Luan, D.%
, Amodei, D.%
\BCBL {}\ \BBA {} Sutskever, I.%
\end{APACrefauthors}%
\unskip\
\newblock
\APACrefYearMonthDay{2019}{}{}.
\newblock
{\BBOQ}\APACrefatitle {Language Models are Unsupervised Multitask Learners}
  {Language models are unsupervised multitask learners}.{\BBCQ}.
\newblock
\begin{APACrefURL} \url{https://api.semanticscholar.org/CorpusID:160025533}
  \end{APACrefURL}
\PrintBackRefs{\CurrentBib}

\bibitem [\protect \citeauthoryear {%
Vaswani%
\ \protect \BOthers {.}}{%
Vaswani%
\ \protect \BOthers {.}}{%
{\protect \APACyear {2017}}%
}]{%
NIPS2017_3f5ee243}
\APACinsertmetastar {%
NIPS2017_3f5ee243}%
\begin{APACrefauthors}%
Vaswani, A.%
, Shazeer, N.%
, Parmar, N.%
, Uszkoreit, J.%
, Jones, L.%
, Gomez, A\BPBI N.%
\BDBL {}Polosukhin, I.%
\end{APACrefauthors}%
\unskip\
\newblock
\APACrefYearMonthDay{2017}{}{}.
\newblock
{\BBOQ}\APACrefatitle {Attention is All you Need} {Attention is all you
  need}.{\BBCQ}
\newblock
\BIn{} I.~Guyon\ \BOthers {.}\ (\BEDS), \APACrefbtitle {Advances in Neural
  Informationyo Systems} {Advances in neural informationyo systems}\
  (\BVOL~30).
\newblock
\APACaddressPublisher{}{Curran Associates, Inc.}
\newblock
\begin{APACrefURL}
  \url{https://proceedings.neurips.cc/paper_files/paper/2017/file/3f5ee243547dee91fbd053c1c4a845aa-Paper.pdf}
  \end{APACrefURL}
\PrintBackRefs{\CurrentBib}

\bibitem [\protect \citeauthoryear {%
Wei%
\ \protect \BOthers {.}}{%
Wei%
\ \protect \BOthers {.}}{%
{\protect \APACyear {2021}}%
}]{%
DBLP:journals/corr/abs-2109-01652}
\APACinsertmetastar {%
DBLP:journals/corr/abs-2109-01652}%
\begin{APACrefauthors}%
Wei, J.%
, Bosma, M.%
, Zhao, V\BPBI Y.%
, Guu, K.%
, Yu, A\BPBI W.%
, Lester, B.%
\BDBL {}Le, Q\BPBI V.%
\end{APACrefauthors}%
\unskip\
\newblock
\APACrefYearMonthDay{2021}{}{}.
\newblock
{\BBOQ}\APACrefatitle {Finetuned Language Models Are Zero-Shot Learners}
  {Finetuned language models are zero-shot learners}.{\BBCQ}
\newblock
\APACjournalVolNumPages{CoRR}{abs/2109.01652}{}{}.
\newblock
\begin{APACrefURL} \url{https://arxiv.org/abs/2109.01652} \end{APACrefURL}
\PrintBackRefs{\CurrentBib}

\bibitem [\protect \citeauthoryear {%
Wolf%
\ \protect \BOthers {.}}{%
Wolf%
\ \protect \BOthers {.}}{%
{\protect \APACyear {2019}}%
}]{%
wolf2019huggingface}
\APACinsertmetastar {%
wolf2019huggingface}%
\begin{APACrefauthors}%
Wolf, T.%
, Debut, L.%
, Sanh, V.%
, Chaumond, J.%
, Delangue, C.%
, Moi, A.%
\BDBL {}others%
\end{APACrefauthors}%
\unskip\
\newblock
\APACrefYearMonthDay{2019}{}{}.
\newblock
{\BBOQ}\APACrefatitle {Huggingface's transformers: State-of-the-art natural
  language processing} {Huggingface's transformers: State-of-the-art natural
  language processing}.{\BBCQ}
\newblock
\APACjournalVolNumPages{arXiv preprint arXiv:1910.03771}{}{}{}.
\PrintBackRefs{\CurrentBib}

\end{thebibliography}
\bibliographystyle{apacite}

\end{document}